\def\year{2022}\relax
\documentclass[letterpaper]{article} 
\usepackage{aaai22}  
\usepackage{times}  
\usepackage{helvet}  
\usepackage{courier}  
\usepackage[hyphens]{url}  
\usepackage{graphicx} 
\usepackage{natbib}  
\usepackage{caption} 
\DeclareCaptionStyle{ruled}{labelfont=normalfont,labelsep=colon,strut=off} 
\usepackage{algorithm}
\usepackage{algorithmic}

\usepackage{newfloat}
\usepackage{listings}
\floatstyle{ruled}
\newfloat{listing}{tb}{lst}{}
\floatname{listing}{Listing}
\title{Explainability in autonomous pedagogically structured scenarios}
\author{
    Minal Suresh Patil
}

\affiliations{
    Umeå universitet\\

    UNIVERSITETSTORGET 4\\
    907 36 Umeå, Sweden\\
    minal.patil@umu.se
}

\usepackage{bibentry}

\begin{document}

\maketitle

\begin{abstract}
In this work, we present the notion of explainability 
for decision-making processes in a pedagogically 
structured autonomous environment. Multi-agent systems that are 
structured pedagogically consist of pedagogical teachers and 
learners that operate in environments in which both are 
sometimes not fully aware of all the states in the environment and beliefs of other 
agents thus making it challenging to explain their 
decisions and actions with one another. This work emphasises the need for robust 
and iterative explanation-based communication 
between the 
pedagogical teacher and the learner. Explaining the rationale 
behind multi-agents decisions in an interactive, partially 
observable environment is necessary to build trustworthy and 
reliable communication between pedagogical teachers and the 
learners. On-going research is primarily focused on 
explanations of the agent’s behaviour towards humans, 
and there is a lack of research in inter-agent explainability.
\end{abstract}

\section{Introduction}
Pedagogy refers to “the interplay between teachers,
students and the learning environment and the learning
tasks” \cite{murphy2012pedagogy}. Pedagogy includes how teachers
and students relate together through instructional methods
implemented in a classroom-like setting. There are two
primary approaches to pedagogy: teacher-centred and
learner-centred. Though at first sight, these two
approaches seem on the opposite side of the spectrum,
however, they often can accompany each other in
achieving objectives. For instance, a teacher-centred
pedagogy is crucial when the teacher is introducing a new
theme and objective to the learners and whilst learner-
centred pedagogy allows for these learners to explore and
gain a deeper understanding of these objectives on their
own. In a learning scenario, the learner-centered approach will allow agents to create new knowledge  by integrating prior knowledge of the environment with unique experiences of its own. 
The pedagogical teacher not only enables this process
but also generates and structures the settings for learning
the environment. Another approach which is a relatively
the new pedagogical method is the learning-centred pedagogy
\cite{bremner2019learner} combines the best of both worlds, where
the pedagogical teacher considers local and contextual
information of the environment, such as the availability of
resources, time and cost. In Multi-Agent Systems (MAS),
the learning-centred approach is more effective since the pedagogical teacher ensures adaptability of their
pedagogical approaches based on the environment. For example, MAS have
become an integral part of the defence industry and play an
active role in various military activities such as path
planning, surveillance, assessment and attack. However, agents' explanation generation in uncertain and dynamic environments is a challenging problem for MAS. Early relevant works have demonstrated the power of explanations \cite{van2004explainable} for combat purposes, which recorded decisions made by agents during a combat mission, provide reasons for the decisions and handle counterfactual queries through a knowledge guided execution as opposed to plans generated by the agents. As pedagogical MAS become more self-supporting and independent of human intervention, explanation generation between agents becomes a pivotal point in understanding their interactions with each other and the environment.

\section{Problem Identification}
Explainable inter-agent pedagogy involves a teacher(s) communicating and assigning tasks to a learner(s), and these agents produce explanations for their decisions during their course of action. Through this work, we define a mandate as instructions communicated and explanations generated by the teacher to the learner and vice-versa in a human-style format. However, understanding the internal representations and beliefs between learners is an essential aspect for inter-agent explainability \cite{omicini2020not}. Thus an explainable pedagogically structured environment must have the following attributes:

\begin{itemize}
\item Learners that can counterfactually think and
communicate with the pedagogical teacher. Counterfactual thinking considers alternative scenarios for past or future events: what might transpire or what might have transpired? This allows learners to gain valuable experience from their mistakes, enabling them to do better in similar tasks in the future. This allows for the learners to rigorously test the environment for the optimal reward through trustworthy and reliable communication with other learners.
\item Open MAS must allow for agents to enter and leave the
environment at any instance without the need for human
intervention through robust and reliable explanations
during a planning or scheduling task.
\item In scenarios where the teachers have complete knowledge of the system, they must have the ability to break down a labyrinth task into a series of simple and explainable sub-tasks under resource, time and cost constraints.

\end{itemize}

\section{Theory of Mind for Learners}

Theory of Mind (ToM) \cite{premack1978does} is
an important concept that involves thinking about mental
states, both of your own and those of others. This enables
humans to infer intentions of others, and also understand
what is going in someone’s mind, such as beliefs,
expectations and fears. The ToM concept is critical in
providing explanations to other agents in terms of beliefs
and expectations and be able to update their beliefs using
optimal explanations. It is rather necessary to adequately
understand inter-agent communication for an effective
collaborative planning task and provide explanations when
necessary. A significant development in the explainability
of agent’s decision and communication ability is to
integrate ToM since it is a well-established framework
when humans collaborate and generate
explanations for their actions whilst achieving their
objectives as well as the team’s objectives.
To be able to measure the trust of an explanation in a multi-
agent environment, a Partially-Observable Markov
decision process (POMDP) was implemented to generate
explanations and measure the trust using survey data
\cite{wang2016impact}. Furthermore, \cite{chakraborti2017plan} proposed a framework where explanations produced by a robot were reconciled to a human’s model.

Thus, we can see it is critical to understand not only trust but also the the technique used to produce the explanation.
Intelligent multi-agent systems are fundamental
elements of an autonomous environment. According to \cite{wooldridge2009introduction}, the purpose of an agent is to achieve a
its objectives in a stochastic environment through
computational processes in a flexible and self-governing
manner without the need for regular human intervention.
Preferably, learners should learn from their experience of the environment in a continuous manner and must be able to
possess the ability to cooperate and interact with each
other. We propose an extension of Wooldridge's work \cite{wooldridge2009introduction} identifying four essential contextual
capabilities of learners in a learning-centred approach environment:

\begin{itemize}
\item \textit{Autonomy:} \\ The ability to perform complex tasks and
subtasks without the need for humans to intervene by
using their own internal beliefs and intentions.
\item \textit{Awareness:} \\ The ability for agents to perceive and be
aware of their environment and take appropriate
decisions to achieve their objectives.
\item \textit{Proactivity:} \\ An agent’s ability to take control of its
actions to achieve its goals.
\item \textit{Sociability:} \\ An agent’s ability to interact with other
agents to understand their beliefs and intentions whilst
achieving their objectives.

\end{itemize}

\section{Beliefs, Desires and Intentions for Learners}

The Beliefs, Desires and Intentions (BDI) model \cite{georgeff1998belief} is a framework commonly used whilst building
intelligent agents. The underlying philosophical argument
for using BDI for MAS is that humans interact with the
environment using the following criteria: a set
of beliefs consist of knowledge of the environment, a set
of desires consist of motivation to complete objectives and
a collection of intentions consist of plans to achieve the agent’s
objectives. Beliefs represent interim knowledge of the
agents and usually is subject to change as and when the
agent explores the world, i.e. environment. The BDI model
uses core concepts that humans use to reason and act in
everyday situations, thus integrating this framework for the
learning-centred approach in MAS is necessary. Desires
of an agent represent the top priority objective that must be
achieved during its lifetime, i.e. a planning episode.
Intentions of agents are resource bound and represent the
commitment to achieving the objective.
According to Torsun \cite{torsun1995foundations}, to achieve goals,
the agents must have the ability to undertake the following;
Commitment; Communication; Conflict Resolution;
Cooperation; Interaction; Negotiation. The need for
coordinated behaviour between the teacher and the learner
and how they share and explain information such as objectives, knowledge, strategies and plans to make decisions and and take necessary actions is important to achieve the common objective. Taking forward the learning-centred approach for an explainable pedagogically structured environment  we define four important
desiderata:

\begin{itemize}

\item\textit{Behavioural Explanations:} \\
Behavioural explanations must be generated by the agents
to explain their behaviour to the teacher as well as to other
agents in the system. These explanations could be
developed through their past experiences whilst planning a
similar task or produce explanations based on agents that
have planned similar tasks. Agents could also provide
contrastive explanations and reasons for their behaviour in
achieving one objective over another to their teacher.

\item \textit{Intention Explanations:} \\
Intention explanations are necessary explanations for
determining the course of action to achieve objectives of
one’s self and the overall goal. Post-hoc explanations for
choosing a certain of action to achieve could provide
insights into how agents prioritise planning tasks on the
fly.

\item \textit{Timely Explanations:} \\
Timely explanations must be provided by the learner to the teacher. It is possible in planning scenarios, where constant monitoring of all the agents is not feasible. Thus, providing timely explanations to teacher regarding what set of instruction are top priority and what set of instructions can be pursued later on during the course of planning.

\item \textit{Perceptual Explanations:} \\
Perceptual explanations are explanations that agents must
be provided once they gather their own as well as the other
agents’ knowledge (beliefs, desires or intentions) of the
environment after the teacher has issued a set of
instructions. Learners must have the ability to explain how
and why it decided to merge or discard knowledge or
strategies from other learners.
\end{itemize}

\section{Challenges and Opportunities}

\textbf{\textit{Trustworthiness}:} Building trust into MAS is critical in planning applications for defence and military purposes. One of the significant challenges of
multi-agent explainable AI Planning is agents
exhibiting malicious behaviour by sharing false plans or
strategies. Often, MAS comprises of asynchronous agents
and heterogeneous agents. In asynchronous MAS, updates
are executed asynchronously whilst maintaining the
overall/global objective and heterogeneous MAS is
comprised of sub-agents or sub-modules which are not
uniformly distributed in the system. The untrustworthiness
of MAS could arise due to internal conflicts, faults,
environmental conditions or when agents are stressed to
achieve a particular objective, thus making agents
untrustworthy of their actions and decisions. Chang and
Kuo proposed a Markov Chain Trust Model \cite{chang2008markov} that
uses the trust level of an agent as a state, and through the
agent’s experience and behaviour, the trust level is
computed.
\vspace{5mm} \\
\noindent\textbf{\textit{Curse of Dimensionality during Belief Augmentation of
Agents}:} In a MAS planning environment, the issue of belief
augmentation occurs when agents interact and exchange
information about the agent’s states. According to Gmytrasiewicz and
Doshi \cite{gmytrasiewicz2005framework} they propose an Interactive-POMDP which
maintains their own belief as well the beliefs about other
agents. This often results in an increase in the complexity of
the belief space due to nesting of beliefs of all the agents,
which often leads to intractability. Partially Observable
Markov Decision Processes (POMDPs) provide a natural
model for sequential decision and planning under
uncertainty. However, finding exact solutions for finite-
horizon POMDPs are PSPACE-complete \cite{papadimitriou1987complexity} and for infinite-horizon POMDPs is
undecidable \cite{madani1999undecidability}. For I-POMDPs the
complexity is at least PSPACE-complete because it can be
accounted for the fact that an I-POMDP can consist of
multiple POMDPs of other learners in the system. As the
size of belief dimension increases, a solution that
accommodates the agent’s beliefs over
all other agents’ actions, intentions and beliefs are still
open to research.

\section{Conclusion}
This work has introduced four important
explainable desiderata which can provide a road-map for
multi-agent systems in a pedagogically structured environment. Overall, establishing a standard metric for
explanations for a pedagogically structured environment is of
paramount importance. In critical domains such as defence
and marine, Situational-Awareness (SA), Context-
Awareness (CA) and Active Perception (AP) describe an agent's awareness and perception and to be able to explain
their actions and decisions during planning are of utmost
importance and almost life-saving. Existing works
from sociology, psychology, and cognitive science can be
leveraged to make pedagogically structured environments explainable, transparent and reliable.


\section{Acknowledgments}
This work was partially supported by the Wallenberg AI, Autonomous
Systems and Software Program (WASP) funded by the Knut and
Alice Wallenberg Foundation. 


\end{document}